\def\year{2019}\relax
\documentclass[letterpaper]{article} 
\usepackage{aaai19}  
\usepackage{times}  
\usepackage{helvet}  
\usepackage{courier}  
\usepackage{url}  
\usepackage{graphicx}  
\usepackage{amsfonts}
\usepackage{algorithm}
\usepackage[noend]{algpseudocode}
\usepackage{amsmath}
\usepackage{multirow}
\usepackage{array}
\usepackage{amssymb}

\def\b{\ensuremath\boldsymbol}

\begin{document}
%
\title{Quantized Fisher Discriminant Analysis}
\author{Benyamin Ghojogh\thanks{The first two authors contributed equally to this work. Also, Sepideh Shaterian Bidgoli helped the authors significantly.}, Ali Saheb Pasand\footnotemark[1], Fakhri Karray, Mark Crowley \\
Department of Electrical and Computer Engineering, University of Waterloo, Waterloo, ON, Canada\\
\{bghojogh, ali.sahebpasand, karray, mcrowley\}@uwaterloo.ca \\
}
\maketitle
\begin{abstract}
This paper proposes a new subspace learning method, named Quantized Fisher Discriminant Analysis (QFDA), which makes use of both machine learning and information theory. There is a lack of literature for combination of machine learning and information theory and this paper tries to tackle this gap. QFDA finds a subspace which discriminates the uniformly quantized images in the Discrete Cosine Transform (DCT) domain at least as well as discrimination of non-quantized images by Fisher Discriminant Analysis (FDA) while the images have been compressed. This helps the user to throw away the original images and keep the compressed images instead without noticeable loss of classification accuracy. We propose a cost function whose minimization can be interpreted as rate-distortion optimization in information theory. We also propose quantized Fisherfaces for facial analysis in QFDA. Our experiments on AT\&T face dataset and Fashion MNIST dataset show the effectiveness of this subspace learning method.
\end{abstract}

\section{Introduction}

Supervised subspace learning is useful for embedding high dimensional data, such as images, in a low dimensional subspace for better separation of classes. Fisher Discriminant Analysis (FDA)\cite{ghojogh2019fisher,friedman2001elements}, first proposed in \cite{fisher1936use}, was one of the first supervised subspace learning methods which is based on scatters and variances of data. The FDA subspace tries to separate the classes from one another while the data instances within every class collapse to a small region \cite{ghojogh2019feature}. 
Recently, deep FDA \cite{diaz2017deep,diaz2019deep} was proposed which uses a least squares approach \cite{ye2007least,zhang2010regularized}.

On the other hand, recently, it was empirically shown that compression can improve the classification accuracy. In \cite{ozah2019compression}, a pre-trained deep neural network was fed with the compressed test images with JPEG (uniform quantization in DCT domain) compression. Note that they work with data in the pixel domain and not frequency domain. They empirically showed that this compression can improve the classification accuracy, opening a question for further theoretical investigations.

In this paper, we propose Quantized FDA (QFDA), which makes a bridge between machine learning \cite{friedman2001elements} and information theory \cite{cover2012elements}.
In QFDA, we want an optimal subspace which separates the classes favorably for the quantized data while minimizing the distortion and entropy (rate) as much as possible, without sacrificing classification performance. 
Our aim is to find a subspace for quantized data which is at least as good as the subspace for non-quantized data in terms of separation of classes. Therefore, one can throw away the non-compressed data with large volume and merely store the quantized data and the learned directions spanning the subspace. 
In other words, we compress the data but the separation of classes is still acceptable. 

In this paper, we use the following notations:
\begin{itemize}
\item $n$: training sample size
\item $n_j$: training sample size in the $j$-th class
\item $d$: dimensionality of data in the input space
\item $d'$: dimensionality of data after zero-padding
\item $c$: number of classes
\item $\b{x}_i$: $i$-th training data instance
\item $\b{x}_i^{(j)}$: $i$-th training data instance in the $j$-th class
\item $\mathbb{R}^{d \times n} \ni \b{X} = [\b{x}_1, \dots, \b{x}_n]$ whose dimensionality becomes $\mathbb{R}^{d' \times n}$ after zero-padding.
\item $\mathbb{R}^{d' \times n_j} \ni \b{X}_j = [\b{x}_1^{(j)}, \dots, \b{x}_n^{(j)}]$ whose dimensionality becomes $\mathbb{R}^{d' \times n_j}$ after zero-padding.
\item $\b{x}_{t,i}$: $i$-th test (out-of-sample) data instance
\item $\mathbb{R}^{d' \times n_t} \ni \b{X}_t = [\b{x}_{t,1}, \dots, \b{x}_{t,n_t}]$ whose dimensionality becomes $\mathbb{R}^{d' \times n_t}$ after zero-padding.
\end{itemize}
Other notations will be introduced at their first appearance in the paper.

This paper is organized as follows. Section \ref{section_FDA} reviews FDA. Uniform quantization in the frequency domain is reviewed in Section \ref{section_JPEG}. The proposed QFDA is explained in Section \ref{section_QFDA}. The connection of QFDA and rate-distortion optimization in the field of compression is discussed in Section \ref{section_rate_distortion_optimization}. Section \ref{section_experments} reports the experiments including the proposed quantized Fisherfaces for face analysis in QFDA. Finally, Section \ref{section_conclusion} concludes the paper and discusses the possible future direction.

\section{Fisher Discriminant Analysis}\label{section_FDA}

Let $\b{U} \in \mathbb{R}^{d \times d}$ be the projection matrix whose columns $\{\b{u}_j\}_{j=1}^d$ are the projection directions spanning the subspace so the subspace is the column-space of $\b{U}$. If we truncate the projection matrix to have $\mathbb{R}^{d \times p} \ni \b{U} = [\b{u}_1, \dots, \b{u}_p]$, the subspace is spanned by $p$ projection directions and it will be a $p$ dimensional subspace where $p \leq d$. 

The projection into the subspace and reconstruction of training and out-of-sample data are \cite{wang2012geometric}:
\begin{align}
&\mathbb{R}^{p \times n} \ni \widetilde{\b{X}} := \b{U}^\top \b{X}, \label{equation_projection_severalPoint} \\
&\mathbb{R}^{d \times n} \ni \widehat{\b{X}} := \b{U}\b{U}^\top \b{X} = \b{U}\widetilde{\b{X}}, \label{equation_reconstruction_severalPoint} \\
&\mathbb{R}^{p \times n_t} \ni \widetilde{\b{X}}_t := \b{U}^\top \b{X}_t, \label{equation_outOfSample_projection_severalPoint} \\
&\mathbb{R}^{d \times n} \ni \widehat{\b{X}}_t := \b{U}\b{U}^\top \b{X}_t = \b{U}\widetilde{\b{X}}_t, \label{equation_outOfSample_reconstruction_severalPoint}
\end{align}
where $\widetilde{\b{X}} = [\widetilde{\b{x}}_1, \dots, \widetilde{\b{x}}_n]$, $\widehat{\b{X}} = [\widehat{\b{x}}_1, \dots, \widehat{\b{x}}_n]$, $\widetilde{\b{X}}_t = [\widetilde{\b{x}}_{t,1}, \dots, \widetilde{\b{x}}_{t,n_t}]$, and $\widehat{\b{X}}_t = [\widehat{\b{x}}_{t,1}, \dots, \widehat{\b{x}}_{t,n_t}]$ are projection and reconstruction of training data and projection and reconstruction of out-of-sample data, respectively.

FDA was first proposed in \cite{fisher1936use}. See recent reviews and analysis in \cite{ghojogh2019fisher,friedman2001elements}. It maximizes the \textit{Fisher criterion} \cite{xu2006analysis,fukunaga2013introduction}:
\begin{align}\label{equation_optimization_FDA_criterion}
&\underset{\b{U}}{\text{maximize}} ~~~ f_F(\b{U}) := \frac{d_B(\b{U})}{d_W(\b{U})} :=  \frac{\textbf{tr}(\b{U}^\top \b{S}_B\, \b{U})}{\textbf{tr}(\b{U}^\top \b{S}_W\, \b{U})}, 
\end{align}
where $\textbf{tr}(.)$ denotes the trace of matrix.
The Fisher criterion $f_F(\b{U})$ is a generalized Rayleigh-Ritz Quotient \cite{parlett1998symmetric}.
Hence, the optimization in Eq. (\ref{equation_optimization_FDA_criterion}) is equivalent to:
\begin{equation}\label{equation_optimization_FDA}
\begin{aligned}
& \underset{\b{U}}{\text{maximize}}
& & \textbf{tr}(\b{U}^\top \b{S}_B\, \b{U}), \\
& \text{subject to}
& & \b{U}^\top \b{S}_W\, \b{U} = \b{I},
\end{aligned}
\end{equation}
where the $\b{S}_B$ and $\b{S}_W$ are the between and within scatters, respectively, defined as:
\begin{align}
&\mathbb{R}^{d \times d} \ni \b{S}_B := \sum_{j=1}^c n_j (\b{\mu}_j - \b{\mu}) (\b{\mu}_j - \b{\mu})^\top, \label{equation_between_scatter} \\
&\mathbb{R}^{d \times d} \ni \b{S}_W := \sum_{j=1}^c \sum_{i=1}^{n_j} (\b{x}_i^{(j)} - \b{\mu}_j) (\b{x}_i^{(j)} - \b{\mu}_j)^\top, \label{equation_within_scatter}
\end{align}
where the mean of the $j$-th class is:
\begin{align}\label{equation_mean_of_class}
\mathbb{R}^{t} \ni \b{\mu}_j := \frac{1}{n_j} \sum_{i=1}^{n_j} \b{x}_i^{(j)}.
\end{align}

The total scatter can be considered as the summation of the between and within scatters \cite{ye2007least,welling2005fisher}:
\begin{align}
\b{S}_T = \b{S}_B + \b{S}_W \implies \b{S}_B = \b{S}_T - \b{S}_W.
\end{align}
Therefore, the Fisher criterion or Eq. (\ref{equation_optimization_FDA_criterion}) can be written as:
\begin{align}\label{equation_optimization_FDA_criterion_with_S_T}
&f_F(\b{U}) = \frac{d_T(\b{U})}{d_W(\b{U})} - 1 := \frac{\textbf{tr}(\b{U}^\top \b{S}_T\, \b{U})}{\textbf{tr}(\b{U}^\top \b{S}_W\, \b{U})} - 1. 
\end{align}
The $-1$ is a constant can be dropped in the optimization problem; therefore, the optimization in FDA can be:
\begin{equation}\label{equation_optimization_FDA_with_S_T}
\begin{aligned}
& \underset{\b{U}}{\text{maximize}}
& & \textbf{tr}(\b{U}^\top \b{S}_T\, \b{U}), \\
& \text{subject to}
& & \b{U}^\top \b{S}_W\, \b{U} = \b{I}.
\end{aligned}
\end{equation}
The Lagrangian of the problem is \cite{boyd2004convex}:
\begin{align*}
\mathcal{L} = \textbf{tr}(\b{U}^\top \b{S}_T\, \b{U}) - \textbf{tr}\big(\b{\Lambda}^\top (\b{U}^\top \b{S}_W\, \b{U} - \b{I})\big),
\end{align*}
where $\b{\Lambda}$ is a diagonal matrix including the Lagrange multipliers. 
Setting the derivative of the Lagrangian to zero gives:
\begin{align}
& \mathbb{R}^{d \times p} \ni \frac{\partial \mathcal{L}}{\partial \b{U}} = 2\b{S}_T \b{U} - 2\b{S}_W\b{U} \b{\Lambda} \overset{\text{set}}{=} \b{0} \nonumber \\
&\implies \b{S}_T\, \b{U} = \b{S}_W\, \b{U} \b{\Lambda}, \label{equation_scatter_generalized_eigendecomposition}
\end{align}
which is the generalized eigenvalue problem $(\b{S}_T, \b{S}_W)$ \cite{ghojogh2019eigenvalue}.
Hence, FDA directions are the eigenvectors in this generalized eigenvalue problem.
Comparing the Eq. (\ref{equation_optimization_FDA_with_S_T}) with the optimization of PCA \cite{jolliffe2011principal,ghojogh2019unsupervised} shows that PCA captures the orthonormal directions with the maximum variance of data; however, the FDA has the same goal but with the orthonormal directions which are manipulated with $\b{S}_W$.  

One of the ways to solve the generalized eigenvalue problem $(\b{S}_T, \b{S}_W)$ is \cite{ghojogh2019eigenvalue}:
\begin{align}
\b{S}_T\, \b{U} = \b{S}_W\, \b{U} \b{\Lambda} &\implies \b{S}_W^{-1} \b{S}_T\, \b{U} = \b{U} \b{\Lambda} \nonumber \\
&\implies \b{U} = \textbf{eig}(\b{S}_W^{-1} \b{S}_T), \label{equation_FDA_dirty_solution}
\end{align}
where $\textbf{eig}(.)$ stacks the eigenvectors column-wise. 
In order to guarantee that $\b{S}_W$ is not singular, we strengthen its diagonal:
\begin{align}\label{equation_FDA_solution_regularized}
\b{U} = \textbf{eig}\big((\b{S}_W + \varepsilon \b{I})^{-1} \b{S}_T\big),
\end{align}
where $\varepsilon$ is a very small positive number, large enough to make $\b{S}_W$ full rank. In the literature, this approach is known as regularized discriminant analysis \cite{friedman1989regularized}.

We can write the total and within scatters in matrix form:
\begin{align}
&\mathbb{R}^{d \times d} \ni \b{S}_T := \b{X} \b{H} \b{X}^\top, \label{equation_total_scatter_matrixForm} \\
&\mathbb{R}^{d \times d} \ni \b{S}_W := \sum_{j=1}^c \b{X}_j\, \b{H}_j\, \b{X}_j^\top, \label{equation_within_scatter_matrixForm}
\end{align}
respectively, where $\mathbb{R}^{n \times n} \ni \b{H} := \b{I}_n - (1/n) \b{1}_n\b{1}_n^\top$ and $\mathbb{R}^{n_j \times n_j} \ni \b{H}_j := \b{I}_{n_j} - (1/n_j) \b{1}_{n_j}\b{1}_{n_j}^\top$ are centering matrices, $\b{I}_n$ is the $n \times n$ identity matrix, and $\b{1}_n$ is the vector of ones with dimensionality $n$. Note that the centering matrix is symmetric and idempotent. 

\section{Uniform Quantization in DCT Domain}\label{section_JPEG}

In this section, we describe a simple version of uniform quantization in the Discrete Cosine Transform (DCT) domain \cite{ahmed1974discrete,rao2014discrete}. This quantization method defines the well-known JPEG (Joint Photographic Experts Group) compression \cite{mitchell1992digital,wallace1992jpeg}.

We first center the data, i.e., remove their mean:
\begin{align}\label{equation_centering_data}
\b{X} \leftarrow \b{X} \b{H}.
\end{align}
Then, we reshape every image $\b{x}$ (a column of $\b{X}$) to its image-array form. The image is then zero-padded to have height and width having integer number of ($8 \times 8$)-pixel blocks \cite{wallace1992jpeg}. Let $d' \geq d$ denote the new dimensionality of image after zero-padding.
Afterwards, the image is divided to its blocks. Let $\b{f}(a,b)$ with $a,b \in \{0, \dots, 7\}$ denote an image block. A two-dimensional DCT transform \cite{watson1994image} is applied to every block \cite{wallace1992jpeg}:
\begin{equation}
\begin{aligned}
\b{F}(\alpha,\beta) &= \frac{1}{4}\, c(\alpha)\, c(\beta) \Big(\sum_{a=0}^7 \sum_{b=0}^7 \b{f}(a,b) \\
&~~~~~\cos \frac{(2 a + 1) \alpha \pi}{16} \cos \frac{(2 b + 1) \beta \pi}{16} \Big), 
\end{aligned}
\end{equation}
where $\b{F}(\alpha,\beta)$ is the signal in the DCT domain with $\alpha, \beta \in \{0, \dots, 7\}$. Therefore, for every block, we have $64\, (=8 \times 8)$ frequencies. The $c(\alpha)$ and $c(\beta)$ are:
\begin{align}
c(\alpha), c(\beta) := 
\left\{
\begin{array}{ll}
  1 / \sqrt{2} & \text{if } \alpha, \beta = 0,\\
  1 & \text{otherwise}.
\end{array}
\right.
\end{align}
We denote the reshaped $\b{F}$ to a vector by $\b{F}' \in \mathbb{R}^{64}$. Moreover, let $\b{F}'_x \in \mathbb{R}^{d'}$ denote the image $\b{x} \in \mathbb{R}^{d'}$ transformed to the DCT (frequency) domain.

Let $q(.)$ denote the uniform quantization function \cite{cover2012elements}:
\begin{align}\label{equation_quantization_map}
\b{F}' \in \mathbb{R}^{64} \mapsto q(\b{F}'; \b{m}) \in \mathbb{R}^{64},
\end{align}
where $\mathbb{R}^{64} \ni \b{m} = [m_0, \dots, m_{63}]^\top$ is the vector whose $k$-th element, $m_k$, is the number of quantization levels for the $k$-th frequency. More specifically, $m_k \in \{2, \dots, \ell_k\}$ where $\ell_k$ is an upperbound on the number of quantization levels.
In order to calculate $\ell_k$, we first bootstrap a sample of $s$ images from the training dataset. This bootstrap is an estimation of the whole dataset according to Monte-Carlo approximation \cite{robert2013monte}. We calculate:
\begin{align}
\ell_k := \text{round}\big(\!\max(|\b{F}'_{i,b}(k)|)\big),
\end{align}
where $i \in \{1, \dots, n\}$, $b \in \{1, \dots, d' / 64\}$, $|.|$ denotes absoulte value, $\b{F}'$ is the reshaped DCT block $\b{F}$ to a vector, and $\b{F}'_{i,b}(k)$ is the $k$-th frequency in the $b$-th block of the $i$-th training image.

\begin{figure}[!t]
\centering
\includegraphics[width=3in]{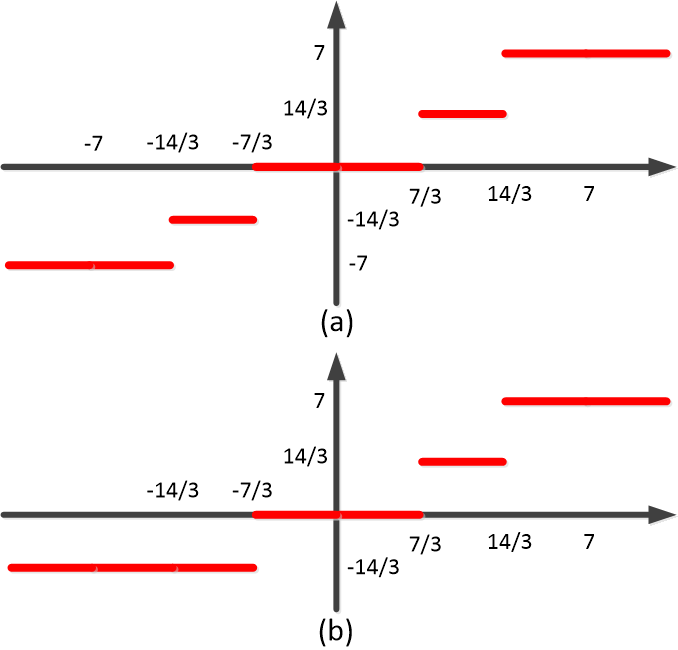}
\caption{Examples for uniform quantization where $\ell_k=7$, $t_1=3$, $t_2=7/3$, and (a) $m_k=5$, (b) $m_k=4$. The horizontal and vertical axes correspond to the input value and quantized value, respectively.}
\label{figure_quantization_example}
\end{figure}

The Eq. (\ref{equation_quantization_map}) quantizes every frequency in every block of the image where the quantization levels for a frequency in different blocks are the same.
The mapping $q$ has the following steps:
\begin{align}
\b{F}'(k) := 
\left\{
\begin{array}{ll}
  \ell_k & \text{if } \b{F}'(k) \geq \ell_k,\\
  -\ell_k & \text{if } \b{F}'(k) \leq -\ell_k,
\end{array}
\right. 
\end{align}
\begin{align}
\text{If } m_k \text{ is even: } \quad
\b{F}'(k) := 
-t_2 & \quad \text{ if } \b{F}'(k) \leq -t_2,
\end{align}
\begin{align}
\b{F}'(k) := \text{sign}\big(\b{F}'(k)\big) \times \frac{\ell_k}{t_1 - 1} \times \Big\lfloor t_1 \times \frac{|\b{F}'(k)|}{\ell_k} \Big\rfloor,
\end{align}
where $\lfloor . \rfloor$ denotes the floor function (rounding to the largest integer less than or equal to the input value), and:
\begin{align}
t_1 := 
\left\{
\begin{array}{ll}
  (m_k + 2) / 2 & \text{if } m_k \text{ is even},\\
  (m_k + 1) / 2 & \text{if } m_k \text{ is odd}.
\end{array}
\right. 
\end{align}
\begin{align}
t_2 := \frac{\ell_k}{t_1} (t_1 - 2).
\end{align}
Two examples of quantization using the above formulae are provided in Fig. \ref{figure_quantization_example}. 
We denote the quantized signal of image $\b{x} \in \mathbb{R}^{d'}$ by $\breve{\b{x}} \in \mathbb{R}^{d'}$. We take the notations $\mathbb{R}^{d' \times n} \ni \breve{\b{X}} = [\breve{\b{x}}_1, \dots, \breve{\b{x}}_n]$ and $\mathbb{R}^{d' \times n_j} \ni \breve{\b{X}}_j = [\breve{\b{x}}_1^{(j)}, \dots, \breve{\b{x}}_n^{(j)}]$.

\section{Quantized Fisher Discriminant Analysis}\label{section_QFDA}

\subsection{Quantized Fisher Criterion}

In QFDA, the total and within scatters are defined as:
\begin{align}
&\mathbb{R}^{d' \times d'} \ni \b{S}_T = \breve{\b{X}} \b{H} \breve{\b{X}}^\top + \lambda \b{X} \b{H} \breve{\b{X}}^\top, \label{equation_QFDA_S_T}\\
&\mathbb{R}^{d' \times d'} \ni \b{S}_W = \sum_{j=1}^c \big(\breve{\b{X}}_j \b{H}_j \breve{\b{X}}_j^\top + \lambda \b{X}_j \b{H}_j \breve{\b{X}}_j^\top\big), \label{equation_QFDA_S_W}
\end{align}
respectively, where $\lambda > 0$ is a hyperparameter controlling the relative importance of the first and second terms. The parameters in $\b{S}_T$ and $\b{S}_W$ can be different but we use the same parameter for the sake of simplicity. 

The first aim of QFDA is to solve the following problem:
\begin{equation}\label{equation_optimization_QFDA_Fisher}
\begin{aligned}
& \underset{\b{U}, \b{m}}{\text{maximize}}
& & f_Q(\b{U}, \b{m}) := \frac{\textbf{tr}(\b{U}^\top \b{S}_T\, \b{U})}{\textbf{tr}(\b{U}^\top \b{S}_W\, \b{U})}, \\
& \text{subject to}
& & m_k \in \{2, 3, \dots, \ell_k\}, ~ \forall k \in \{0, \dots, 63\}.
\end{aligned}
\end{equation}
We name the objective function in Eq. (\ref{equation_optimization_QFDA_Fisher}) the \textit{quantized Fisher criterion} and denote it by $f_Q(\b{U}, m)$.

\subsection{Minimization of Rate}

At the same time, QFDA desires to minimize the rate of the quantized data for better compression. 
The rate of quantization of the $k$-th frequency is defined as \cite{cover2012elements}:
\begin{align}\label{equation_rate}
r_k := \sum_{t=1}^{m_k} p_t\, l_t,
\end{align}
where $l_t$ is the length of the $t$-th quantization level and $p_t$ is the area under the Probability Density Function (PDF) of the DCT signal, $\b{F}(k)$, in the interval of the $t$-th quantization level:
\begin{align}
p_t := \int_{\b{F}(k) \in \text{interval } t} \mathtt{f}_k\big(\b{F}(k)\big)\,\, d\b{F}(k),
\end{align}
where $\mathtt{f}(\b{F}(k))$ is the PDF of the DCT transformed signal for the $k$-th frequency. 
In order to estimate this PDF, we first take the same bootstrapped images which we have from Section \ref{section_JPEG}. In other words, we use the $k$-th frequency in all the blocks of the sampled images. Then, for every frequency, we use kernel density estimation \cite{scott2015multivariate} with Gaussian kernel because Gaussian is the most common distribution in the real-world signals, also supported by central limit theorem \cite{hazewinkel2001central}. 
Note that, again, the estimation of PDF with a bootstrapped sample from data follows the Monte-Carlo approximation \cite{robert2013monte}.

The rate can be approximated by the entropy because if we use $l_t \approx -\log_2(p_t)$ in Eq. (\ref{equation_rate}), we have:
\begin{align}\label{equation_rate_entropy}
r_k \approx - \sum_{t=1}^{m_k} p_t\, \log_2(p_t),
\end{align}
which is the summation of entropies in the quantization intervals. In other words, if the frequency has a lot of information because of significant changes amongst the blocks/images, it should have a large number of quantization levels and we expect that its rate/entropy to be large. 
We use the approximation in Eq. (\ref{equation_rate_entropy}) for the rate in QFDA optimization. 
We calculate an overall rate for the dataset by averaging over the rates of $64$ frequencies:
\begin{align}
\bar{r} := \frac{1}{64} \sum_{k=0}^{63} r_k.
\end{align}


\subsection{Optimization for QFDA}

The complete cost function in QFDA is:
\begin{equation}\label{equation_optimization_QFDA_cost}
\begin{aligned}
& \underset{\b{m}}{\text{minimize}}
& & -f_Q(\b{U}, \b{m}) + \gamma\, \bar{r}, \\
& \text{subject to}
& & m_k \in \{2, 3, \dots, \ell_k\}, ~ \forall k \in \{0, \dots, 63\},
\end{aligned}
\end{equation}
which minimizes the rate but maximizes the quantized Fisher criterion where the optimization variable is the vector containing the number of quantization levels for the $64$ frequencies. The $\gamma > 0$ is the regularization parameter in this optimization. 

The columns of $\b{U} \in \mathbb{R}^{d' \times d'}$ are the \textit{quantized Fisher directions}. We can truncate this matrix to have a $p$-dimensional quantized Fisher subspace where $p \leq d'$. In other words, the column space of $\mathbb{R}^{d' \times d'} \ni \b{U} = [\b{u}_1, \dots, \b{u}_p]$ is the \textit{quantized Fisher subspace}.
We calculate the quantized Fisher directions by solving Eq. (\ref{equation_optimization_QFDA_Fisher}) for a given $\b{m}$. The solution is similar to the solution of Eq. (\ref{equation_optimization_FDA_with_S_T}) which is Eq. (\ref{equation_FDA_solution_regularized}) where we use Eqs. (\ref{equation_QFDA_S_T}) and (\ref{equation_QFDA_S_W}) for $\b{S}_T$ and $\b{S}_W$, respectively. 

We solve the Eq. (\ref{equation_optimization_QFDA_cost}) using Particle Swarm Optimization (PSO) \cite{kennedy2010particle} which is a powerful metaheuristic optimization method \cite{burke2005search}. 
The reason that we solve this optimization problem using a heuristic method is that this cost is a little ugly in a mathematical sense and also the search space is discrete which makes the problem harder. 
The cost of PSO is the cost in Eq. (\ref{equation_optimization_QFDA_cost}) where before feeding a solution (particle) $\b{m}$ to it, we first project the input to the valid set of values. We denote the projection for the $k$-th frequency by $\Pi(m_k)$ and define it as:
\begin{align}
\Pi(m_k) := 
\left\{
\begin{array}{ll}
  \ell_k & \text{if } m_k > \ell_k, \\
  2 & \text{if } m_k < 2, \\
  \lceil m_k - 0.5 \rceil & \text{if } 2 \leq m_k \leq \ell_k.
\end{array}
\right.
\end{align}
Note that the quantized Fisher criterion in the cost is calculated as explained before for a given $\b{m}$ (particle). The rate, existing in the cost, is also calculated using Eq. (\ref{equation_rate_entropy}).

\section{Connection to Rate-Distortion Optimization}\label{section_rate_distortion_optimization}

Quantization and compression usually deal with the rate-distortion optimization \cite{ortega1998rate,cover2012elements}, where the rate and distortion are in trade-off meaning that lower distortion usually comes with higher rate and vice versa. In the rate-distortion optimization, the rate is minimized for better compression while the distortion is also tried to be minimized for better preserved quality.

The optimization in Eq. (\ref{equation_optimization_QFDA_cost}) can be interpreted as the rate-distortion optimization. We explain the reason in the following. 
The criterion in Eq. (\ref{equation_optimization_QFDA_Fisher}) is a generalized Rayleigh-Ritz Quotient \cite{parlett1998symmetric}.
Hence, the optimization (\ref{equation_optimization_QFDA_Fisher}), for a given $\b{m}$, is equivalent to:
\begin{equation}\label{equation_optimization_QFDA_Fisher_2}
\begin{aligned}
& \underset{\b{U}}{\text{minimize}}
& & -\textbf{tr}(\b{U}^\top \b{S}_T\, \b{U}), \\
& \text{subject to}
& & \b{U}^\top \b{S}_W\, \b{U} = \b{I}, 
\end{aligned}
\end{equation}
where maximization has been changed to minimization by negating the objective function.
The Lagrangian relaxation \cite{boyd2004convex} of the problem is:
\begin{equation}\label{equation_optimization_QFDA_Fisher_2_Lagrange_relaxation}
\begin{aligned}
\underset{\b{U}}{\text{minimize}}~ \mathcal{L} := &-\textbf{tr}(\b{U}^\top \b{S}_T\, \b{U}) \\
& + \textbf{tr}\big(\b{\Lambda}^\top (\b{U}^\top \b{S}_W\, \b{U} - \b{I})\big),
\end{aligned}
\end{equation}
where the diagonal of $\b{\Lambda}$ are the Lagrange multipliers.
Now, consider the Eq. (\ref{equation_optimization_QFDA_cost}) whose Lagrange relaxation is similarly as the following:
\begin{equation}\label{equation_optimization_QFDA_cost_Lagrange_relaxation}
\begin{aligned}
\underset{\b{m}}{\text{minimize}}~ \mathcal{L} := &-\textbf{tr}(\b{U}^\top \b{S}_T\, \b{U}) \\
& + \textbf{tr}\big(\b{\Lambda}^\top (\b{U}^\top \b{S}_W\, \b{U} - \b{I})\big) + \gamma\, \bar{r},
\end{aligned}
\end{equation}
where we have dropped the terms concerning the constraints in Eq. (\ref{equation_optimization_QFDA_cost}) for simpler analysis. 
This equation shows that we are minimizing the within scatter of the projected data. According to Eq. (\ref{equation_QFDA_S_W}), the terms $\breve{\b{X}}_j \b{H}_j \breve{\b{X}}_j^\top$ and $\b{X}_j \b{H}_j \breve{\b{X}}_j^\top$ are minimized. Minimization of the former minimizes the cloud of $j$-th quantized class in the QFDA subspace. Therefore, it plays the role of minimization of the entropy (or information) which helps minimization of the rate. 
On the other hand, minimization of $\b{X}_j \b{H}_j \breve{\b{X}}_j^\top$ is like minimization of the distortion.
The reason is that this term shows the variance (dissimilarity) of the quantized (i.e., distorted) data from non-distorted data. Hence, its minimization plays the minimization of distortion.

Although the terms $\breve{\b{X}} \b{H} \breve{\b{X}}^\top$ and $\b{X} \b{H} \breve{\b{X}}^\top$ also exist in $\b{S}_T$ which is maximized in Eq. (\ref{equation_optimization_QFDA_cost_Lagrange_relaxation}), but notice that they represent the total variance of data which guarantees better separation of classes in the subspace. Therefore, it makes more sense to consider $\b{S}_W$ for rate-distortion optimization and not $\b{S}_T$. Moreover, the rate $\bar{r}$ exists in Eq. (\ref{equation_optimization_QFDA_cost_Lagrange_relaxation}) which is minimized as expected. Therefore, in that equation, rate is minimized (in $\bar{r}$ and $\b{S}_W$) and distortion is also minimized (in $\b{S}_W$) as we have in rate-distortion optimization. 
It is also noteworthy that, according to Eqs. (\ref{equation_QFDA_S_W}) and (\ref{equation_optimization_QFDA_cost_Lagrange_relaxation}), the trade-off of minimization of rate and distortion is controlled by the two hyperparameters $\lambda$ and $\gamma$. 

\begin{table}[!t]
\label{table_validation_ATT}
\renewcommand{\arraystretch}{1.3}  
\centering
\scalebox{0.6}{    
\begin{tabular}{l || c | c | c | c | c}
\hline
\hline
 & 0.1 & 0.5 & 1 & 1.5 & 2 \\
\hline
\hline
0.01 & 0.122 $\pm$ 0.062 & 0.129 $\pm$ 0.049 & 0.127 $\pm$ 0.052 & 0.182 $\pm$ 0.069 & 0.146 $\pm$ 0.058 \\
\hline
0.1 & 0.117 $\pm$ 0.074 & 0.135 $\pm$ 0.053 & 0.165 $\pm$ 0.074 & 0.157 $\pm$ 0.064 & 0.121 $\pm$ 0.063 \\
\hline
1 & 0.125 $\pm$ 0.064 & \textbf{0.107 $\pm$ 0.066} & 0.169 $\pm$ 0.068 & 0.132 $\pm$ 0.063 & 0.153 $\pm$ 0.063 \\
\hline
10 & 0.123 $\pm$ 0.054 & 0.117 $\pm$ 0.056 & 0.117 $\pm$ 0.049 & 0.156 $\pm$ 0.060 & 0.151 $\pm$ 0.079 \\
\hline
\hline
\end{tabular}%
}
\caption{The average error of 10-NN classification of AT\&T glasses dataset for the validation data projected to the QFDA subspace. The rows and columns correspond to the values for $\gamma$ and $\lambda$, respectively.}
\end{table}

\begin{table}[!t]
\label{table_validation_Fashion_MNIST}
\renewcommand{\arraystretch}{1.3}  
\centering
\scalebox{0.6}{    
\begin{tabular}{l || c | c | c | c | c}
\hline
\hline
 & 0.1 & 0.5 & 1 & 1.5 & 2 \\
\hline
\hline
0.01 & 0.266 $\pm$ 0.125 & 0.266 $\pm$ 0.126 & 0.256 $\pm$ 0.130 & 0.265 $\pm$ 0.130 & 0.263 $\pm$ 0.122 \\
\hline
0.1 & 0.266 $\pm$ 0.125 & 0.266 $\pm$ 0.126 & 0.256 $\pm$ 0.130 & 0.265 $\pm$ 0.130 & 0.263 $\pm$ 0.122 \\
\hline
1 & 0.266 $\pm$ 0.125 & 0.266 $\pm$ 0.126 & 0.256 $\pm$ 0.130 & 0.265 $\pm$ 0.130 & \textbf{0.263 $\pm$ 0.122} \\
\hline
10 & 0.266 $\pm$ 0.125 & 0.266 $\pm$ 0.126 & 0.256 $\pm$ 0.130 & 0.265 $\pm$ 0.130 & 0.263 $\pm$ 0.122 \\
\hline
\hline
\end{tabular}%
}
\caption{The average error of 10-NN classification of Fashion MNIST dataset for the validation data projected to the QFDA subspace. The rows and columns correspond to the values for $\gamma$ and $\lambda$, respectively.}
\end{table}

\section{Experiments}\label{section_experments}

\subsection{Quantized Fisherfaces vs. Fisherfaces}

We used the AT\&T face dataset \cite{web_att_face_dataset} which includes $400$ facial images of $40$ different people. The images have different expressions and poses making this dataset hard enough.  
The size of images, in this dataset, are $112 \times 92$ pixels. For computational reasons, we resampled the images to $56 \times 46$ pixels because calculation of eigenvectors takes time and every iteration of optimization includes calculation of eigenvectors.
We divided the images into two classes of having and not having eye glasses. 
We split data to training, test, and validation sets with proportions $60\%$, $20\%$, and $20\%$, respectively.

\begin{figure}[!t]
\centering
\includegraphics[width=3in]{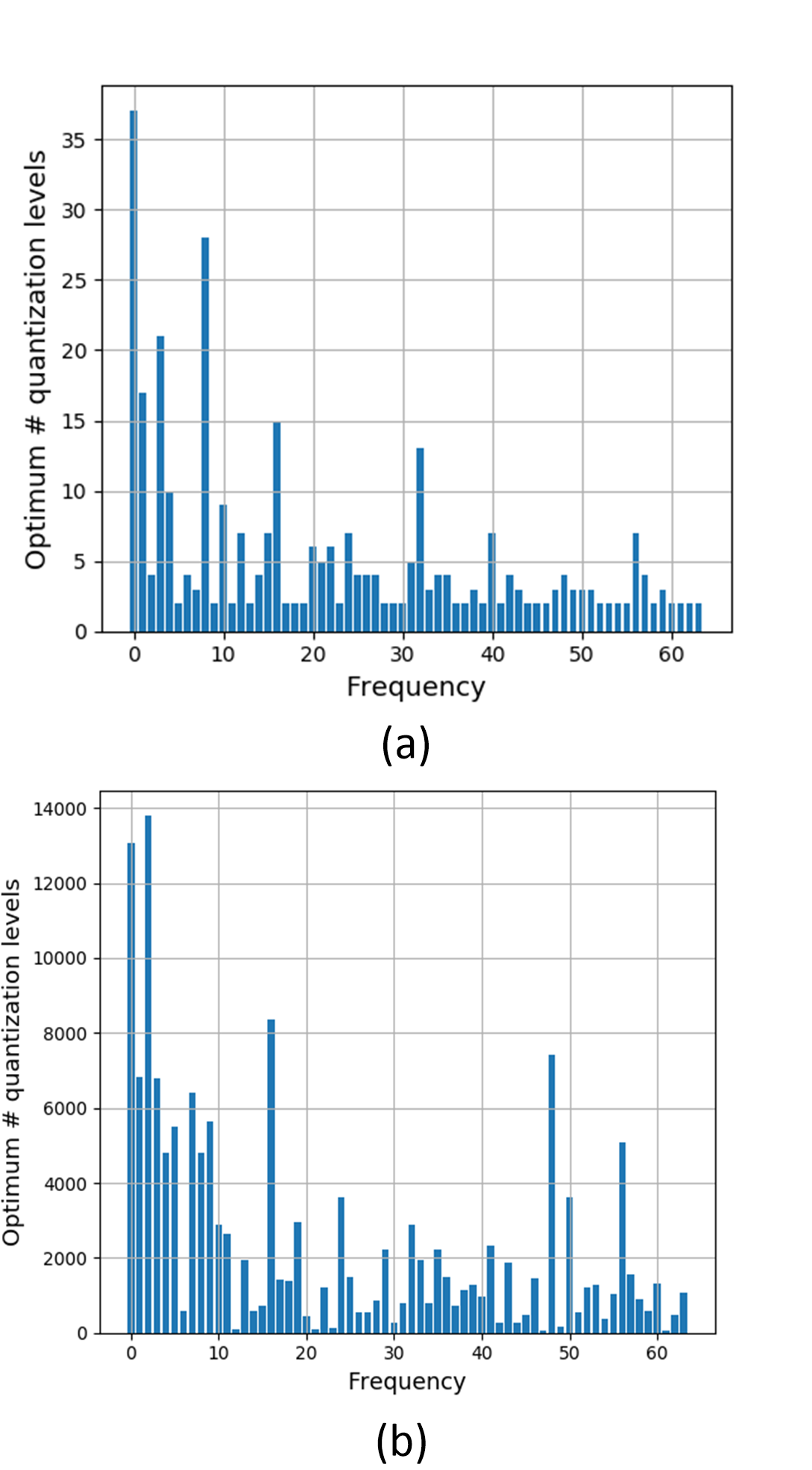}
\caption{The optimum $m$ (number of quantization levels) for the $64$ frequencies in (a) AT\&T glasses dataset and (b) Fashion MNIST dataset.}
\label{figure_optimum_values}
\end{figure}

\begin{figure*}[!t]
\centering
\includegraphics[width=6.8in]{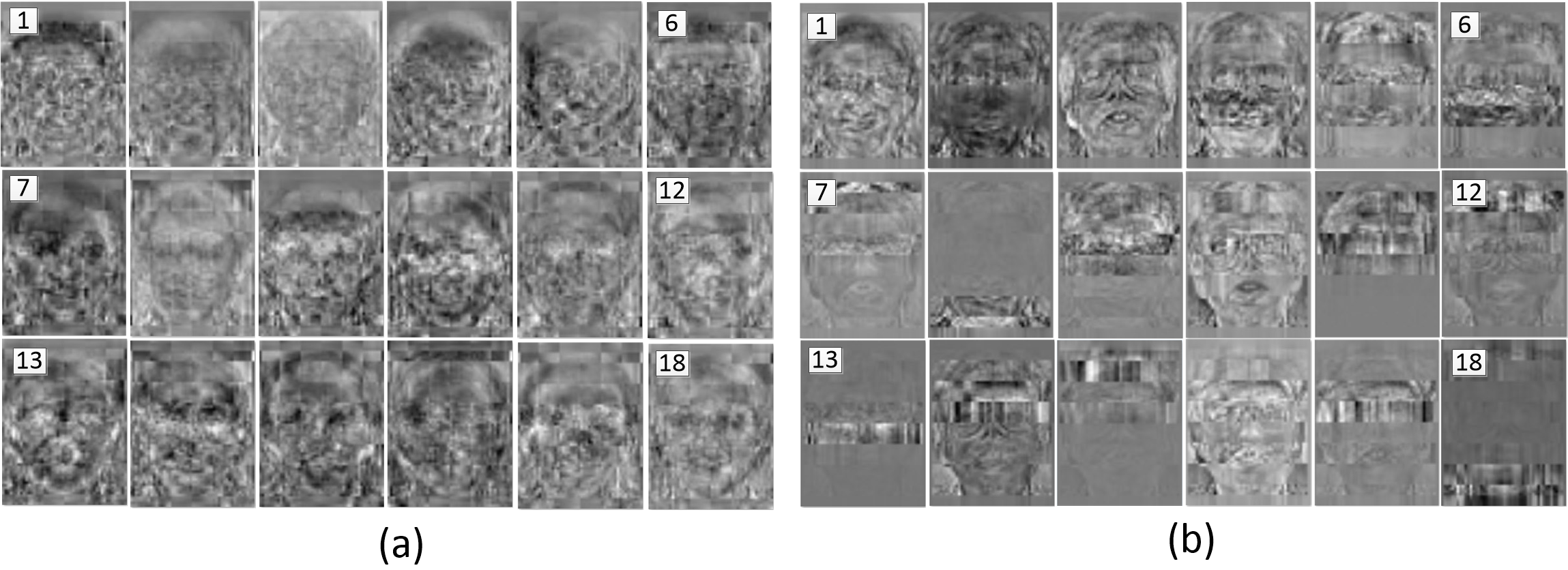}
\caption{The inverse DCT transform of eighteen leading ghost faces: (a) quantized Fisherfaces, (b) Fisherfaces.}
\label{figure_eigen_att}
\end{figure*}

\begin{figure*}[!t]
\centering
\includegraphics[width=5.5in]{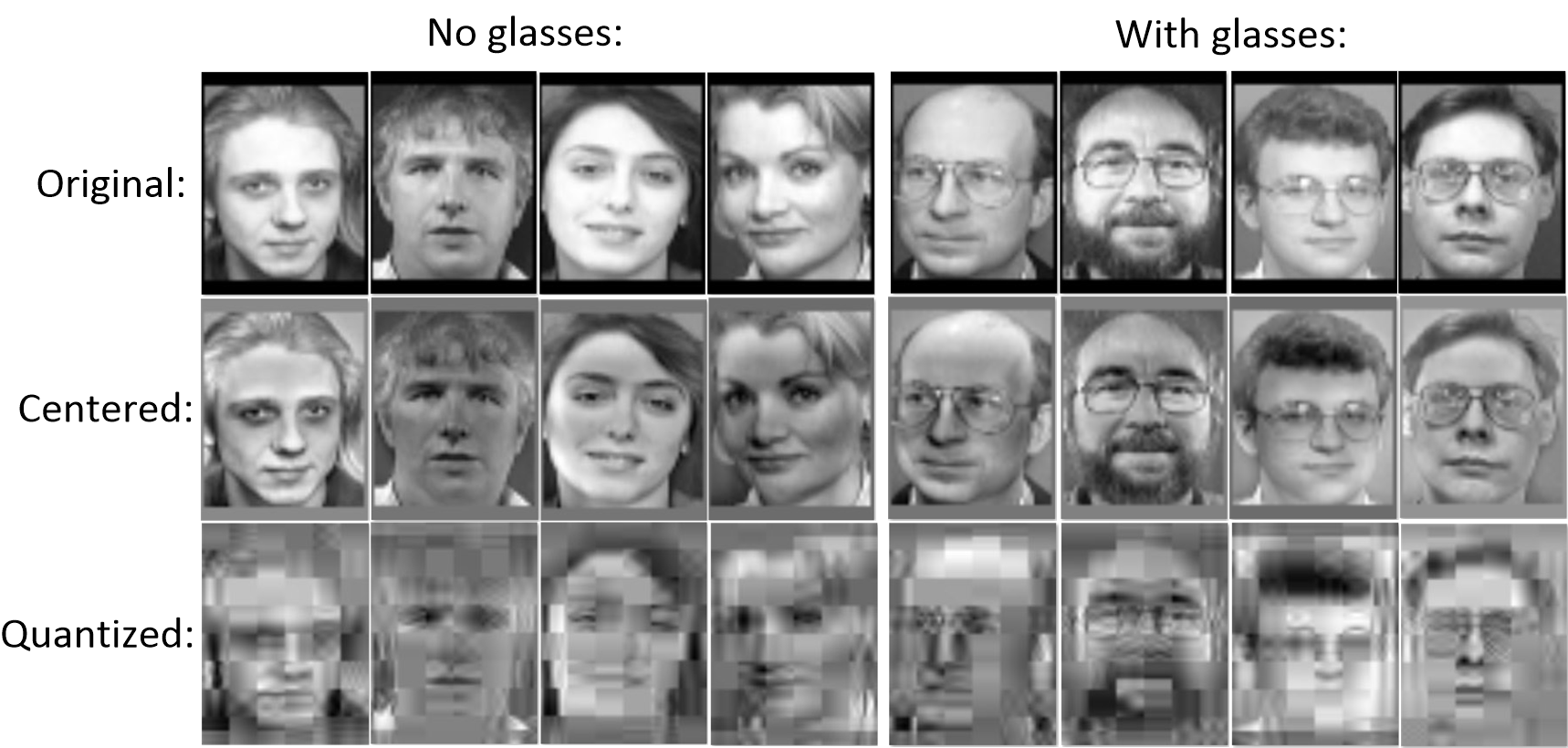}
\caption{The quantized versus original (non-distorted) images in AT\&T glasses dataset where the quantization is done for the optimum $m$ in QFDA.}
\label{figure_quantized_att}
\end{figure*}

We used the validation set for finding the best values for $\gamma$ and $\lambda$. For several permutations of values of these two hyperparameters, we ran the PSO optimization. For every PSO optimization, we used five particles and we found ten iterations to be sufficient. 
A 10-Nearest Neighbor (10-NN) classification was used for evaluating the QFDA on all training, test, and validation sets. Nearest neighbor classification is useful to evaluate the structure of the embedded data in the subspace. 
Note that the training set was used for classification of test and validation sets. 

\begin{figure*}[!t]
\centering
\includegraphics[width=6.5in]{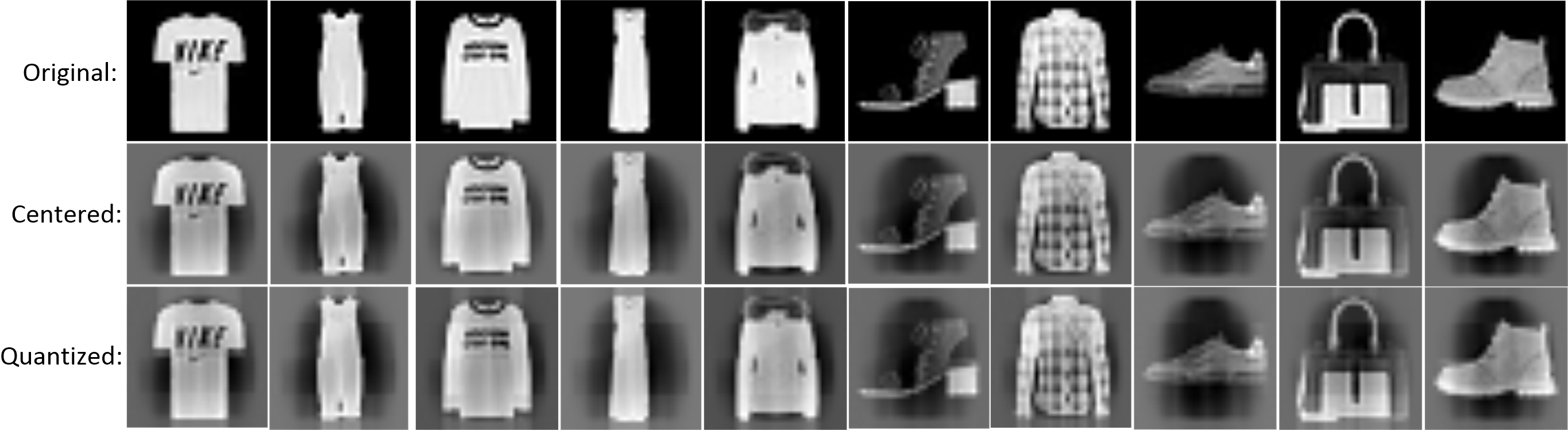}
\caption{The quantized versus original (non-distorted) images in Fashion MNIST dataset where the quantization is done for the optimum $m$ in QFDA.}
\label{figure_quantized_Fashion_MNIST}
\end{figure*}

\begin{figure*}[!t]
\centering
\includegraphics[width=6.8in]{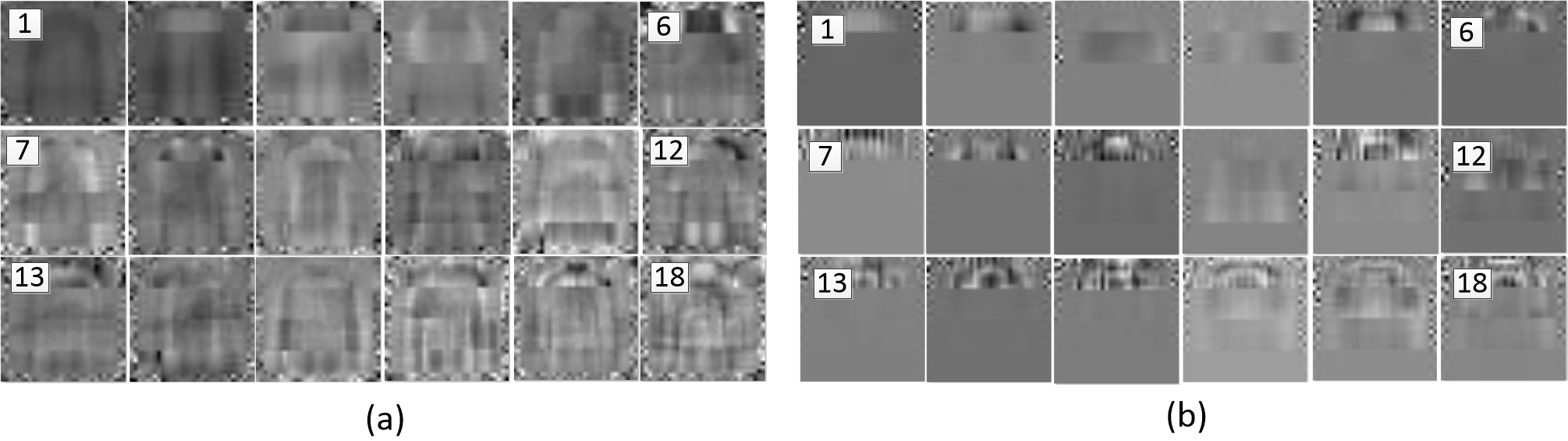}
\caption{The inverse DCT transform of eighteen leading eigenvectors in: (a) QFDA, (b) FDA.}
\label{figure_eigen_Fashion_MNIST}
\end{figure*}

For evaluations, we did the classification considering the first up to the first $20$ dimensions of the subspace. We report the results as the average (and standard deviation) over the $20$ error rates.
Table \ref{table_validation_ATT} reports the validation errors where $\gamma=1$, $\lambda=0.5$ were found to be the best.  
In the following, we report the training, test, and validation errors for these optimum hyperparameters:
\begin{itemize}
\item training: 0.126 $\pm$ 0.032
\item test: 0.208 $\pm$ 0.041
\item validation: 0.107 $\pm$ 0.066
\end{itemize}
Note that the above results are the QFDA results for embedding the quantized DCT transformed images.
The average error rates for embedding the original (non-quantized) DCT transformed images using the FDA:
\begin{itemize}
\item training: 0.112 $\pm$ 0.017
\item test: 0.185 $\pm$ 0.021
\item validation: 0.179 $\pm$ 0.025
\end{itemize}
Except for the validation set, the results of QFDA are slightly worse than FDA; however, we should note that the compression was remarkably high from 20.2 kilo bytes per image to 10 kilo bytes. The drop of classification rate was not significant however. 
The above comparison shows that we can throw away the original data and its FDA subspace and keep the compressed images with the QFDA subspace to classify the quantized classes.

The found optimum number of quantization levels for the $64$ frequencies are depicted as a bar plot in Fig. \ref{figure_optimum_values}-(a). 
The result is convincing because the DC component (frequency $0$) usually has the smallest variance in the pixel domain; thus it has largest variance in the frequency domain. The opposite analysis exists for the highest frequency. Hence, we expect that more quantization levels should be assigned to the lower frequencies.

The inverse DCT transform of the leading eigenvectors (directions) of QFDA and FDA subspaces are shown in Fig. \ref{figure_eigen_att}. The facial eigenvectors, or ghost faces, of FDA are referred to as Fisherfaces in the literature \cite{belhumeur1997eigenfaces}. 
Similarly, we name the ghost faces in QFDA as \textit{quantized Fisherfaces}.
Comparing the Fisherfaces and quantized Fisherfaces shows that quantized Fisherfaces take care of the JPEG blocking resulted from the quantization. Both Fisherfaces and quantized Fisherfaces are capturing the features regarding the eye regions of faces which is expected be3cause the two classes are different in terms of having or not having eye glasses. Note that we have more than one $(=c-1)$ eigenvector here for both FDA and QFDA because we have regularized it with $\varepsilon = 10^{-7}$ in Eq. (\ref{equation_FDA_solution_regularized}).

Recall that in Eq. (\ref{equation_centering_data}), we centered the data. Several zero-padded images, centered data, and quantized images are shown in Fig. \ref{figure_quantized_att}. As mentioned before, the size of every images has been halved without any significant drop in accuracy of classification. As shown in this figure, the eyes and glasses have been quantized with higher quality as expected because the eyes are important for classification between having and not having eye glasses. On the other hand, some non-important facial regions are quantized with less quality which is expected for minimization of rate.

\subsection{Experiments on Fashion MNIST Dataset}

In order to do experiments for more than two classes, we used the Fashion MNIST dataset \cite{web_fashion_mnist_dataset} which includes $10$ classes of different clothing items. The size of images, in this dataset, is $28 \times 28$ pixels. The settings of PSO and cross validation were the same as the previous experiment. Table \ref{table_validation_Fashion_MNIST} shows the validation results for this dataset. This table shows that for this dataset, QFDA is almost robust to the changes of $\gamma$. The reason might be because the size of images is small in this dataset and thus the rate is usually high (distortion is small) because we do not have huge number of dummy pixels. The quantized images, shown in Fig. \ref{figure_quantized_Fashion_MNIST}, also show that distortion is not significant in this dataset, although some level of distortion is observed if one looks at the images carefully.

The table \ref{table_validation_Fashion_MNIST} shows that $\gamma=1, \lambda=2$ can be a good choice for the parameters. 
In the following, we report the training, test, and validation errors for these optimum hyperparameters:
\begin{itemize}
\item training: 0.236 $\pm$ 0.102
\item test: 0.283 $\pm$ 0.133
\item validation: 0.263 $\pm$ 0.122
\end{itemize}
The above results are the QFDA results for embedding the quantized DCT transformed images.
The average error rates for embedding the original (non-quantized) DCT transformed images using the FDA:
\begin{itemize}
\item training: 0.273 $\pm$ 0.086
\item test: 0.315 $\pm$ 0.114
\item validation: 0.317 $\pm$ 0.121
\end{itemize}
This shows that, on this dataset, QFDA has performed much better than FDA. The size of image before and after quantization is almost 7.91 and 7.89 kilo bytes which means that the compression was not significant as explained before.

The found optimum number of quantization levels for the $64$ frequencies are depicted as a bar plot in Fig. \ref{figure_optimum_values}-(b). The same interpretation as before exists for why DC component should have larger number of quantization levels.

The inverse DCT transform of eighteen leading eigenvectors of QFDA and FDA are illustrated in Fig. \ref{figure_eigen_Fashion_MNIST}. This figure shows that the eigenvectors of QFDA have captured more features in comparison to the eigenvectors of FDA. 
The eigenvectors of FDA have captured partially scanned features rather than complete features.
This explains the better results of QFDA.

\section{Conclusion and Future Direction}\label{section_conclusion}

This paper proposed Quantized Fisher Discriminant Analysis (QFDA) which made a bridge between machine learning, manifold learning, and information theory. There is a huge lack of literature for combination of machine learning and information theory and this paper tried to tackle this gap. This method optimized a proposed cost function using PSO algorithm. This cost function can be interpreted as a rate-distortion cost which is used in compression purposes. The quantized Fisherfaces method was also proposed for facial analysis in QFDA. The experiments reporting validation, optimum number of quantization levels, visualization of QFDA eigenvectors, and display of quantized images showed the merit of this new subspace learning method. 

In this paper, we worked on uniform quantization in the DCT domain which is what JPEG compression does.
A possible future work is to consider a general non-uniform quantization. In that case, the variables for quantization optimization will be an integer number of levels, $m_k$ float values for start of quantization intervals, and $m_k$ float values for the mapping values in quantization. Therefore, if the upperbound on $m_k$ is $\ell_k$, the vector of solution (particle in PSO) should have $2\ell_k + 1$ or just $2\ell_k$ dimensions where some of its entries will be zero if $m_k < \ell_k$.

\section*{Acknowledgment}

The authors thank Sepideh Shaterian Bidgoli for her very helpful discussions in developing the idea of this paper.

\bibliographystyle{aaai}
\bibliography{references}

\end{document}